\documentclass{uai2023} 
\usepackage[american]{babel}
\usepackage{natbib} 
\usepackage{mathtools} 
\usepackage{booktabs} 
\usepackage{tikz} 

\usepackage{algorithm}
\usepackage{algorithmic}
\usepackage{amsfonts}
\usepackage{amsmath}
\usepackage{setspace}
\usepackage{relsize}
\usepackage{xcolor}
\usepackage{float}
\usepackage[caption=false,font=footnotesize]{subfig}
\usepackage{newfloat}
\usepackage{listings}
\usepackage{multirow}
\usepackage{booktabs}
\usepackage{float}

\title{PolicyClusterGCN: Identifying Efficient Clusters  for  \\ Training Graph Convolutional Networks}

\author[1]{Saket Gurukar\thanks{Work done when the author was a graduate student at OSU.}}
\author[1]{Shaileshh Bojja Venkatakrishnan}
\author[2,3]{Balaraman Ravindran}
\author[1]{Srinivasan Parthasarathy}
\affil[1]{%
    The Ohio State University, Columbus, Ohio, USA
}
\affil[2]{%
    Indian Institute of Technology Madras, Chennai, India
}
\affil[3]{%
    Robert Bosch Centre for Data Science and AI, Chennai, India
  }

\begin{document}
\maketitle

\begin{abstract}
  Graph convolutional networks (GCNs) have achieved huge success in several machine learning (ML) tasks on graph-structured data. Recently, several sampling techniques have been proposed for the efficient training of GCNs and to improve the performance of GCNs on ML tasks. Specifically, the subgraph-based sampling approaches such as ClusterGCN and GraphSAINT have achieved state-of-the-art performance on the node classification tasks. These subgraph-based sampling approaches rely on {\it heuristics} -- such as graph partitioning via edge cuts --  to identify clusters that are then treated as minibatches during GCN training. In this work, we hypothesize that rather than relying on such heuristics, one can learn a reinforcement learning (RL) policy to compute efficient clusters that lead to effective GCN performance.  To that end, we propose PolicyClusterGCN, an online RL framework that can identify good clusters for GCN training. We develop a novel Markov Decision Process (MDP) formulation that allows the policy network to predict ``importance" weights on the edges which are then utilized by a clustering algorithm (Graclus) to compute the clusters. We train the policy network using a standard policy gradient algorithm where the rewards are computed from the classification accuracies while training GCN using clusters given by the policy. Experiments on six real-world datasets and several synthetic datasets show that PolicyClusterGCN outperforms existing state-of-the-art models on node classification task. 
\end{abstract}

\section{Introduction}

Graph convolution networks (GCNs) learn high-quality node representations of graph-structured data. Such representations allow GCN to achieve state-of-the-art performance on several graph-based machine learning tasks such as node classification \citep{perozzi2014deepwalk}, link prediction \citep{kipf2016variational}, 
molecular graph generation \citep{you2018graph}, 
and recommendation systems \citep{ying2018graph}. However, GCN cannot easily operate on large-scale graphs as it requires $O(nfl)$ memory \citep{chiang2019cluster} where $n, f,$ and $l$ are the number of nodes, features, and GCN layers, respectively.

\begin{figure*}[t]
    \centering
    \includegraphics[width=0.92\linewidth]{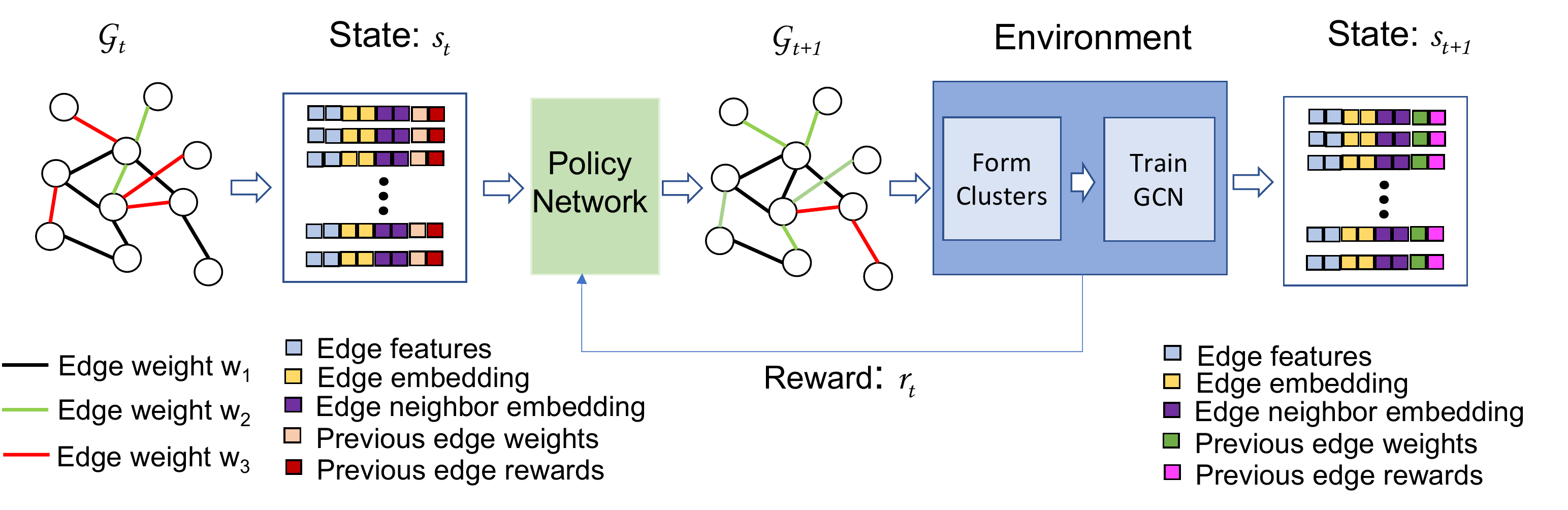}
    \caption{Overview of PolicyClusterGCN (best viewed in color)
    }
    \label{fig:PolicyClusterGCN}
\end{figure*}


To improve the efficiency and scaling of GCNs, several sampling-based models have been proposed recently \citep{liu2021sampling}. Liu et al. categorize these sampling methods as node-wise, layer-wise, and subgraph-based  sampling models. The recent subgraph-based  sampling models such as ClusterGCN \citep{chiang2019cluster} and GraphSAINT \citep{zeng2019graphsaint} have shown superior performance on node classification tasks as compared to GCNs and other sampling models. These subgraph-based models identify subgraphs (clusters) that are treated as minibatches for GCN training.  The resultant mini-batch training helps improve the performance (such as node classification) of the GCN and also helps in scaling GCNs to large graphs.  

Subgraph-based based sampling models rely on {\it predetermined heuristics} -- such as graph partitioning via edge cuts, or communication volume minimization -- to identify subgraphs (clusters). For instance, ClusterGCN \citep{chiang2019cluster} and DistDGL \citep{zheng2020distdgl}  relies on the clustering algorithm, Metis \citep{karypis1997metis} to identify clusters while Aligraph \citep{zhu2019aligraph} relies on four graph partitioning algorithms to identify clusters. GraphSAINT \citep{zeng2019graphsaint}, on the other hand, relies on random subgraph sampling techniques to identify subgraphs. These techniques achieve better performance than GCN by either avoiding the neighborhood expansion problem \citep{chiang2019cluster} in GCN training or by mitigating the high variance in GCN training through normalization \citep{zeng2019graphsaint}.

To the best of our knowledge, such heuristics
are oblivious to impact of formed clusters on GCN training effectiveness. The optimization objective of GCN \citep{kipf2016semi}
suggests that its training performance through clusters is dependent on both the graph structure of the clusters and the distribution of labels within the clusters. 
However, it is difficult to know {\it apriori} which heuristics will lead to the identification of efficient clusters for a given input graph. Here, a cluster configuration is efficient if it results in good GCN  performance.


Identifying efficient clusters is a hard problem: for $k$ clusters, there exists $k^{n}$ possible cluster configurations where $n$ is the number of nodes. Training GCN on $k^{n}$ cluster configurations to find the best performance is not computationally feasible. Hence, we require systematic exploration of the possible configurations. To that end, we propose PolicyClusterGCN, an online reinforcement learning framework that relies on a novel Markov Decision Process (MDP) formulation for computing clusters. The MDP policy is parameterized with a neural network. However, using RL approach to compute efficient clusters is a challenging problem. For instance, a straightforward approach of designing a policy that directly assigns nodes to clusters would lead to clusters with a large fraction of nodes without an immediate edge between them. 
Hence, we design our policy neural network such that it predicts  an ``importance" weight for each edge. Edges that are deemed unimportant are assigned a low weight and vice-versa by the policy. 
A standard edge-cut-based graph partitioning algorithm would then output different cluster configurations based on the edge weights of the graph. This novel setup allows the exploration of a diverse set of clusters while including nodes that are immediately connected in the same clusters.  We train the policy network using a standard policy gradient algorithm where the rewards are computed from the classification accuracies while training GCN using clusters given by the policy.
Our contributions can be summarized as follows: 
\begin{enumerate}
    \item We discover that the choice of clusters has a significant impact on GCN performance.
    \item To compute efficient clusters for GCN training, we formulate a novel MDP in which
    policy predicts edge weights that are utilized by a clustering algorithm to compute clusters.
    \item Our experiments show that GCN trained on clusters identified by PolicyClusterGCN outperforms existing methods on five real-world datasets. 
    \item We also analyze the clusters computed by PolicyClusterGCN by studying their graph structure (via synthetic datasets) and label distribution (via label entropy metric \citep{chiang2019cluster}).
    
    
\end{enumerate}


\section{Notations}

\textbf{Notations}: Let $G(V,E)$ be the input graph $G$ with $|V|$ and $|E|$ number of nodes and edges, respectively. Each edge $e=(u,v,w)$ consists of nodes $u$ and $v$ and has edge weight $w \in \mathbb{Z}^+$ that denotes the strength of connection. 
Let $A \in \mathbb{R}^{|V| \times |V|}$ and $\hat{A} \in \mathbb{R}^{|V| \times |V|}$ be the adjacency matrix and normalized adjacency matrix of graph $G$, respectively.   Let $F \in \mathbb{R}^{|V| \times f}$ be the node features. Let $Z^{(l)}$ be the node embedding at layer $l$ with $Z^{0}=F$ and $W^{(l)}$ be the learning parameters of GCN. The node labels are denoted by $y_L$. 
Let $\mathcal{T_G}$ be the training graph consisting of all the training nodes and edges incident on those training nodes. Similarly, let $\mathcal{V_G}$ and $\mathcal{T}e_\mathcal{G}$ be the validation and test graphs.

\section{Methodology}


\subsection{MDP formulation}

The overview of PolicyClusterGCN is shown in Figure \ref{fig:PolicyClusterGCN}. At a step $t$ in MDP, the agent selects an action that updates all the edge weights of the graph in state $s_t$. As a result, the MDP transitions to new state $s_{t+1}$. 
Our MDP is nonepisodic that continuously trains during each step.
We experimented with episodic MDP formulations, however, we empirically found our presented MDP design to work better.

\begin{algorithm}[t]
\caption{PolicyClusterGCN}
\setstretch{1.05}
\label{alg:policyclustergcn}
\textbf{Input}: $\mathcal{T_G}$ : train graph, $\mathcal{V_G}$ : validation graph, $k$ : num clusters, iters: clustergcn iterations, $\alpha$ : set of discrete edge weights, $y_{T_{L}}$ and $y_{V_{L}}$: train and validation node labels, respectively. \\
\textbf{Parameters}:   $\gamma$, $\theta$, $w$, $\alpha^\theta$, $\alpha^w$, $\epsilon$, $\epsilon_{start}, \epsilon_{end}, \epsilon_{decay}$ \\
\textbf{Output}: Cluster configuration. \\
\begin{algorithmic}[1] 
\FOR{t = 0, 1, 2, ... $T$}
\STATE Assign edge weight of edge $i$  where $1\le i \le |E|$ \\
$w_i=  \begin{cases}
  \pi_\theta(s_{t})[i] & \text{with } 1-\epsilon \text{ prob.} \\
  random (\alpha) & \text{with } \epsilon \text{ prob.}
\end{cases}$
\STATE Decay $\epsilon$ at each step
\STATE Update graph $(\mathcal{T_{G^{'}}})$ with new edge weights. 
\STATE Identify $k$ clusters from $(\mathcal{T_{G^{'}}})$. Change edge weight of $k$ clusters to that of $\mathcal{T_{G}}$
\STATE Train ClusterGCN in Algorithm \ref{alg:ClusterGCN} on the identified clusters of $\mathcal{T_G}$.
\STATE Compute validation score from the above trained ClusterGCN on  $\mathcal{V_G}$.
\IF{validation-score $>$ best-validation-score:}   
\STATE cluster\_configuration = identified clusters
\STATE best-validation-score = validation-score
\ENDIF
\STATE Compute score of edge i = (u,v) as $sc_i = sc_u + sc_v $ \\
    $sc_u=  \begin{cases}
  +1 & \text{if node label predicted correctly }  \\
  -1 & \text{otherwise}
\end{cases}$
\STATE $r_t = \frac{1}{|E|} \sum sc_i$ where $1\le i \le |E|$
\STATE Update parameters $\theta$ and $w$ using equations \ref{eqn:adv} and \ref{eqn:ac}.
\ENDFOR
\RETURN cluster\_configuration
\end{algorithmic}
\end{algorithm}
\smallskip
\noindent \textbf{Agent}: Let $\mathcal{S}$ and $\mathcal{A}$ be the state space and action space, respectively. In this work, a state represents the graph. Let $\alpha$ be the set of possible discrete edge weights on an edge and $a = (e_1, \ldots, e_{|E|})$ be the selected edge weights where $e_i \in \alpha$.
The edge placement policy can then be defined as a mapping $\pi: \mathcal{S} \rightarrow \mathcal{A}$ that assigns an edge weight to all the edges where $\mathcal{A} = \alpha^{|E|}$. 
The goal of the policy network is to find a placement policy $\pi$ that results in efficient clusters.  

\smallskip
\noindent \textbf{Environment}: The environment accepts a graph and its edge weights computed by the agent as its input. It then partitions the graph into several clusters based on the edge weights and trains the GCN model on each cluster.

\smallskip
\noindent \textbf{State}: A state observation $s \in S$ comprises of training graph $\mathcal{T_G}$ with the following features on each edge $e=(u,v)$ : (1) concatenation of node features of nodes $u$ and $v$; (2) concatenation of node embeddings of nodes $u$ and $v$. Here, we utilize an unsupervised graph representation learning method (UGRL), node2vec \citep{grover2016node2vec} to learn the node embeddings (later, we show in the experiments section, that one can also choose other UGRL methods for this step); 
(3) concatenation of embeddings of 1-hop neighbor nodes of nodes $u$ and $v$. A node's neighbor embeddings are aggregated with sum operation. We choose the sum operation as it allows us to capture the graph neighborhood in an expressive manner \citep{xu2018powerful}; (4) the previous $m$ number of edge weights taken by the policy $\pi$; (5) the previous $m$ number of rewards received from the environment.
At the initial state $s_0$, the previous $m$ edge weights and previous $m$ edge rewards are assigned a value of 1 and 0, respectively.

\smallskip
\noindent \textbf{Action}: An action at step $t$ is given by the policy $\pi$: $a_t =\pi(s_t) = (e_1, \ldots, e_{|E|})$. To accelerate the exploration phase through actions, we utilize exponential edge weights. In exponential edge weights, if there are $p$ number of possible actions, an action $i$ would correspond to edge weight $2^{i}$ where $0 \le i \le p$. An edge with $2^p$ weight would have the strongest connection and would be less likely to be eliminated during graph partitioning. The choice of discrete edge weights instead of continuous edge weights for the actions is primarily due to the restriction-induced by graph partitioning algorithms (e.g. Metis \citep{karypis1997metis}, Graclus \citep{dhillon2007weighted}, and MLR-MCL \citep{satuluri2009scalable}).

\begin{algorithm}[t]
\setstretch{1.25}
\caption{ClusterGCN}
\label{alg:ClusterGCN}
\textbf{Input}: $\mathcal{T_{G}}_1, \mathcal{T_{G}}_2, ... \mathcal{T_{G}}_k$ ; iters, $l$
\begin{algorithmic}[1] 
\STATE Let $\mathcal{B}=\{\mathcal{T_{G}}_1, \mathcal{T_{G}}_2, ...\mathcal{T_{G}}_k\}$
\FOR{iter = 1, ... $\le$ iters}
\STATE $\mathcal{B} \leftarrow$  Permute $\mathcal{B}$ in an uniform random manner.
\FOR{x in $\mathcal{B}$}
\STATE $Z_x^{(l+1)} = \sigma(\hat{A}_{\mathcal{G}x}Z_x^{(l)}W^{(l)})$ where $Z_x^{(0)}=F_x$
\STATE minimize $\mathcal{L} = \frac{1}{|y_L|} \mathlarger{\sum}_{i \in y_L} loss (y_i, Z_{x,i}^L)$
\ENDFOR
\ENDFOR
\end{algorithmic}
\end{algorithm}

\begin{table*}[t]
\centering
\resizebox{1.0\linewidth}{!}{
\begin{tabular}{lrrrrc|l}
\toprule
\textbf{Datasets} & \textbf{Nodes}                   & \textbf{Edges}                    &  \textbf{Feats} & \textbf{L/C} & \textbf{train/ val/ test} & \textbf{References} \\
\midrule
Romania           & 41,773                                               & 996,404                                               & 16                                                                                   & 84        & 0.60 / 0.10 / 0.30 &   \multirow{3}{10cm}{\citep{rozemberczki2019gemsec,wang2019inductive, rozemberczki2020little,yang2021boosting, bianconi2021message, zhou2022influence, auletta2019maximizing, bouyer2022influence}}                       \\
Hungary          & 47,538                                               & 445,774                                               & 16                                                                                   & 84         & 0.60 / 0.10 / 0.30 &                                  \\
Croatia          & 54,573                                               & 251,652                                               & 16                                                                                   & 84            & 0.60 / 0.10 / 0.30        &                        \\
\midrule
Twitter         & 2,403  &  37,154 & 9,073  &   73      &   0.55 / 0.20 / 0.25 &  \citep{he2018discovering, he2018clustering, yao2021fuzzy} \\
\midrule
Facebook         &  4,039         &  	88,234         &  	1,283         &      84     & 0.60 / 0.20 / 0.20 &  \citep{he2018discovering, hu2021incorporating,tang2022detection} \\
\midrule
Blogcatalog          & 5,196  & 343,486  & 8189  &   6           & 0.55 / 0.20 / 0.25  & \citep{huang2019graph, huang2018exploring, huang2017accelerated, huang2017label} \\
\bottomrule
\end{tabular}
}
\caption{Dataset statistics. Feats denotes number of features. L/C denotes number of labels or classes. 
}
\label{tab:datasets}
\end{table*}

\smallskip
\noindent \textbf{Reward}: Now, to identify a good cluster configuration from the space of all possible cluster configurations, we require a reward signal about the goodness of a given cluster configuration. 
Here, we treat the performance (such as node classification) of GCN on training nodes as the reward signal by training GCN for a few iterations ($iters$) over the computed clusters. We train GCN for a small number of iterations (for example, 50 iterations in our experiments) to get rewards in a short amount of time. Once the GCN model is trained, we compute a score for each edge $e=(u,v)$ of the training graph by summing scores of nodes $u$ and $v$. Here, a score for a node is +1 if the node label is predicted correctly by the trained GCN while -1 otherwise. In the case of multilabel classification, we compute the score for each label and then sum the scores.
We compute the reward as the mean of all the scores across all the edges.


\subsection{Policy Network Architecture}
PolicyClusterGCN learns the good cluster configuration of a graph by parameterizing the MDP policy using a neural network. Here, one could design the policy network using GCN. However, GCN has a huge memory requirement for large graphs. Hence, we select a two-layer neural network as our policy network. The input to the network is the state of edge $i$ and the output of the network is the predicted edge weight. This design allows the parallelization of the edge weight prediction task. This policy network design can also perform the edge weight prediction task for large graphs.  
The policy network is trained with a standard policy gradient algorithm. 


\subsection{Training}
PolicyClusterGCN is trained with a standard policy gradient algorithm - actor-critic \citep{konda1999actor}. 
Let $\theta$ and $w$ be the policy parameters (actor) and state-value function parameters (critic), respectively. 
Let $r_t$ be the received reward obtained by following policy $\pi_\theta$ at step $t$. 
Then the actor and critic parameters are updated as follows:
\begin{equation}
\label{eqn:adv}
    \delta \leftarrow r_t + \gamma \hat{v} (s_{t+1}, w) - \hat{v}(s_t, w),
\end{equation}
where $\gamma$ is the discount factor and $\hat{v}$ is the value function and,
\begin{equation}
\label{eqn:ac}
\begin{aligned}
w \leftarrow & \textnormal { } w + \alpha^w \delta  \textnormal { }\nabla_w \textnormal { }  \hat{v} (s_t,w)  \\ 
\theta \leftarrow & \textnormal { } \theta + \alpha^\theta \delta  \textnormal { } \nabla_\theta \textnormal { ln } \pi (a_t|s_t, \theta),
\end{aligned}
\end{equation}
    %

 where $\alpha^w > 0$ and $\alpha^\theta > 0$ are the step sizes for actor and critic parameters and $\mathcal{A}_t$ denotes the actions taken by step $t$. 
 
 The training algorithm of PolicyClusterGCN is presented in Algorithm \ref{alg:policyclustergcn}. 
 In each step, we update the edge weights of all the edges.
 Here, given the state $s_{t}$  at step $t$,  we employ an $\epsilon$-greedy method for exploring the edge weights.  We set the edge weight of an edge $i$ to one of the possible edge weights in a uniform random manner with $\epsilon$ probability and set the edge-weight of the edge given by the policy network ($\pi_\theta(s_{t})[i]$) with $1 - \epsilon$ probability. We decay the $\epsilon$ values as $\epsilon_{end} + (\epsilon_{start} - \epsilon_{end}) \times \exp(-1 \times t/\epsilon_{decay}))$. We update the training graph $\mathcal{T_G}$ with the predicted edge weights. 
 
The change in the edge weights of the graph allows us to explore different cluster configurations. We identify clusters using a multi-level graph partitioning algorithm, Graclus \citep{dhillon2007weighted} that allows the identification of imbalanced clusters. Metis \citep{karypis1997metis}, on the other hand, has a load balance constraint that allows limited exploration of diverse cluster configurations. Once, we identify the clusters, we change the edge weight of the non-cut edges to that of the original graph. This change is important as we are interested in identifying good cluster configuration rather than modifying the input graph. 

Next, to compute the reward signal we train the ClusterGCN model (shared in Algorithm \ref{alg:ClusterGCN}). 
In ClusterGCN, given $k$ clusters, we form $\mathcal{T_{G}}_1, \mathcal{T_{G}}_2, ... \mathcal{T_{G}}_k$ subgraphs. Let $\hat{A}_{\mathcal{G}x} \in \mathbb{R}^{n_x \times n_x}$ be the normalized adjacency matrix of subgraph $x$ with $n_x$ number of nodes. Let $Z_x^{(l)} \in \mathbb{R}^{n_x \times f}$ be the node embedding at layer $l$ with $Z_x^{0}=F_x$ where $F_x$ is the initial node features. While $W^{(l)}$ are the learnable parameters.  Then, we train GCN by following the training procedure outlined in ClusterGCN \citep{chiang2019cluster}. The ClusterGCN training process also includes a parameter ($bsize$) to enable a stochastic multi-partition approach and we follow the same approach in our implementation (approach \citep{chiang2019cluster}, not shown in Algorithm \ref{alg:ClusterGCN} for expository simplicity). 
Once we train ClusterGCN, we compute the reward by following the procedure mentioned in ``MDP formulation" section. We also compute the validation score on the supervised task (such as node classification) and store the clusters that resulted in the best validation score.  The policy and critic networks are then trained using equations \ref{eqn:adv} and \ref{eqn:ac}.

\section{Experiments}

\begin{table*}[t]
\centering
\resizebox{0.95\linewidth}{!}{
\begin{tabular}{l|l|cccccc}
\toprule
\textbf{Sampling Strategy}                            &                   \textbf{Model}          & \multicolumn{1}{l}{\textbf{Croatia}} & \multicolumn{1}{l}{\textbf{Romania}} & \multicolumn{1}{l}{\textbf{Hungary}} & \textbf{Facebook}         & \textbf{Twitter}             & \textbf{Blogcatalog}      \\
\midrule
  & GCN & 0.343 & 0.340 & 0.384 & 0.500 & 0.150 & 0.940\\
 & &  \small{($\pm$ 0.030)} & \small{($\pm$ 0.034)} & \small{($\pm$ 0.036)} & \small{($\pm$ 0.031)}  & \small{($\pm$ 0.015)} &  \small{($\pm$ 0.012)} \\ 
 \midrule 
\multirow{2}{*}{Node wise sampling} & GraphSage & 0.355 & 0.352 & 0.399 & 0.478 & 0.132 & 0.933 \\
& &  \small{($\pm$0.011)} & \small{($\pm$0.017)} & \small{($\pm$0.024)} & \small{($\pm$0.022)}  & \small{($\pm$0.022)} &  \small{($\pm$0.005)} \\ [0.1cm] 
 & VR-GCN & 0.338 & 0.366 & 0.396 & 0.525 & 0.156 & 0.953 \\
 & &  \small{($\pm$ 0.026)} & \small{($\pm$  0.030)} & \small{($\pm$  0.030)} & \small{($\pm$ 0.010)}  & \small{($\pm$  0.055)} &  \small{($\pm$ 0.010)} \\
 \midrule 
\multirow{2}{*}{Layer wise sampling} & LADIES & 0.403 & 0.377 & 0.422 & 0.459 & 0.137 & 0.889\\
& & \small{($\pm$ 0.040)} & \small{($\pm$ 0.043)} & \small{($\pm$ 0.043)} & \small{($\pm$0.024)}  & \small{($\pm$0.023)} &  \small{($\pm$0.006)} \\ [0.1cm] 
 & FAST-GCN & 0.404 & 0.375 & 0.383 & 0.368 & 0.115 & 0.832 \\ 
 & &  \small{($\pm$ 0.075)} & \small{($\pm$ 0.070)} & \small{($\pm$ 0.056)} & \small{($\pm$0.009)}  & \small{($\pm$0.034)} &  \small{($\pm$0.032)} \\
 \midrule
\multirow{3}{*}{Subgraph sampling} & RippleWalk & 0.359 & 0.358 & 0.406 & 0.466 & 0.144 & 0.877 \\
& &  \small{($\pm$ 0.027)} & \small{($\pm$ 0.030)} & \small{($\pm$ 0.0340)} & \small{($\pm$ 0.046)}  & \small{($\pm$ 0.052)} &  \small{($\pm$ 0.011)} \\ [0.1cm] 
 & GraphSaint & 0.459 & 0.467 & 0.485 & 0.537 & 0.164 & {\bf 0.959 }\\
  & &  \small{($\pm$0.005)} & \small{($\pm$0.002)} & \small{($\pm$0.001)} & \small{($\pm$0.007)}  & \small{($\pm$0.014)} &  \small{($\pm$0.005)} \\ [0.1cm] 
 & ClusterGCN & 0.403 & 0.387 & 0.420 & 0.510 & 0.148 & 0.950\\
 & &  \small{($\pm$0.012)} & \small{($\pm$0.005)} & \small{($\pm$0.043)} & \small{($\pm$0.006)}  & \small{($\pm$0.032)} &  \small{($\pm$0.004)} \\ [0.1cm] 
 & ClusterHOGCN & 0.453 & 0.469 & 0.488 & 0.557 & 0.171 & 0.951 \\
 & &  \small{($\pm$0.003)} & \small{($\pm$0.006)} & \small{($\pm$0.001)} & \small{($\pm$0.013)}  & \small{($\pm$0.022)} &  \small{($\pm$0.004)} \\
 \midrule
 \multirow{2}{*}{Proposed} & PolicyClusterGCN & 0.428 & 0.413 & 0.453 & 0.515 & 0.155 & 0.952 \\
 & &  \small{($\pm$0.009)} & \small{($\pm$0.001)} & \small{($\pm$0.007)} & \small{($\pm$0.008)}  & \small{($\pm$0.024)} &  \small{($\pm$0.006)} \\ [0.1cm] 
 &PolicyClusterHOGCN & \textbf{0.463} & \textbf{0.478} & \textbf{0.497} & \textbf{0.563} & \textbf{0.189} & 0.956 \\ 
  & &  \small{($\pm$0.003)} & \small{($\pm$0.003)} & \small{($\pm$0.001)} & \small{($\pm$0.001)}  & \small{($\pm$0.016)} &  \small{($\pm$0.006)} \\
\bottomrule
\end{tabular}
}
\caption{Test Micro-f1 score on the node classification task. Results averaged over five independent runs for all the models. 
}
\label{tab:node_classi}
\end{table*}
\subsection{Datasets} We evaluate PolicyClusterGCN on six frequently evaluated datasets. The statistics of these datasets are presented in Table \ref{tab:datasets}. The column train/val/test in Table \ref{tab:datasets} refers to the size of training, validation, and test splits of the dataset.  The  Romania, Hungary, and Croatia datasets were introduced in \citep{rozemberczki2019gemsec} and extensively used in the literature. We evaluate models on multilabel classification tasks on Romania, Hungary, Croatia, Twitter, and Facebook datasets and multiclass classification on the Blogcatalog dataset. 
In the case of Twitter and Facebook datasets, certain labels are present in only a few nodes, hence in our evaluation, we consider only those labels that are present in at least 10 nodes.\textcolor{blue}{} Node attributes are not present for Romania, Hungary, and Croatia datasets. Hence, we factorize the adjacency matrix of these datasets and treat the 16 left singular vectors as node attributes \citep{cukierski2011graph}.

\subsection{Baselines} We consider state-of-the-art GCN training models including several models that perform efficient sampling for GCN training \citep{liu2021sampling}. 
\begin{itemize}
\item \textbf{GCN} \citep{kipf2016semi} : Initial graph convolutional network.
    \item \textbf{GraphSAGE} \citep{hamilton2017inductive} : A node sampling based GCN that samples few l-hop neighbors of nodes to learn the node embeddings. 
    \item \textbf{VR-GCN} \citep{chen2018stochastic}: A node sampling based GCN that utilizies historial activations to sample a smaller number of node's neighbors. 
    \item \textbf{FAST-GCN} \citep{chen2018fastgcn}: A layer sampling based GCN that samples new node neighbors at each layer of GCN.
    \item \textbf{LADIES} \citep{zou2019layer}: A layer sampling based GCN that utilizes importance sampling to sample neighbor-dependent nodes at each layer of GCN.
    \item \textbf{RippleWalk} \citep{bai2021ripple}: A subgraph sampling based GCN that propose a random process to form a subgraph for GCN training.
    \item \textbf{GraphSaint} \citep{zeng2019graphsaint}: A subgraph sampling based GCN that proposed samplers for forming subgraph and introduce normalization techniques to eliminate bias and reduce variance.
    \item \textbf{ClusterGCN} \citep{chiang2019cluster}: A subgraph sampling based GCN that partitions graphs into multiple clusters and treat each cluster as minibatch for GCN training.
    \item \textbf{PolicyClusterGCN}: Our proposed algorithm that performs efficient training of GCN.
    \item \textbf{PolicyClusterHOGCN}: Our proposed algorithm where instead of GCN we train Higher Order GCN (HOGCN) model \citep{zeng2019graphsaint}. Each layer in HOGCN combines GraphSAGE-concat \citep{hamilton2017inductive} layer with MixHop \citep{abu2019mixhop} layer. We tune the same HOGCN parameters for both GraphSaint and PolicyClusterHOGCN.
\end{itemize}

\subsection{Training Details} 
PolicyClusterGCN is implemented in PyTorch.  For baselines we utilize the code provided by the authors.
For PolicyClusterGCN
we apply Graclus \citep{dhillon2007weighted} partitioning method to formulate clusters. 

We set the number of GCN layers to 2 for all the baselines. We tune the following hyper-parameters for all the baselines: learning rates = [0.01, 0.001], dropouts = [0.0, 0.1, 0.2] and embedding dimensions = [128, 512, 1024]. The number of epochs is set to 1500 for all the models.  For GraphSAINT, we also tune the normalization parameters =  [``norm", ``norm-nn"] and embedding aggregation process = [``mean", ``concatenation"]. For ClusterGCN and PolicyClusterGCN, we tune the number of clusters = [4, 8, 16, 32] and number of clusters for each batch = [1, 4, 8]. For all the models, we perform the evaluation on the same five train/val/test splits and report the average test micro-f1 score over five runs. For PolicyClusterGCN, we set the number of previous edge weights and the number of the previous edge rewards value $m$ to 5. 
To get the node embeddings, we set the node2vec parameters as walk-length=40, context-size=5, walks-per-node=20, p=1.0, q=1.0. During development, we played with the parameters of node2vec and found that PolicyClusterGCN is not sensitive to those parameters.  We train PolicyClusterGCN on the cluster configuration identified
 by Algorithm \ref{alg:policyclustergcn}. 

 PolicyClusterGCN parameters: We set the $eps\_decay$ value to 100. The step sizes of $\alpha^w = 0.001$ and $\alpha^\theta = 0.001$. The discount factor $\gamma$ is set to 0.95. During development, we played with the following node2vec parameters: walk-length=[20, 40, 80], context-sizes=[3,5], walks-per-node=[20, 40]. However, we did not find PolicyClusterGCN to be sensitive with respect to node2vec parameters.


All the experiments are conducted on a machine with an NVIDIA Tesla V100 (32 GB memory), an Intel Xeon E5-2680 CPU (28 cores, 2.40GHz), and 128 GB of RAM.

\section{Results}
\subsection{Performance}
We compare the node classification performance of PolicyClusterGCN with other baselines in Table \ref{tab:node_classi}. The values in the parenthesis in Table \ref{tab:node_classi} represents standard deviation. We rely on the Micro-f1 metric for evaluation as it is a frequently used metric in the literature. We observe that our proposed PolicyClusterHOGCN outperforms state-of-the-art models on five out of six real-world datasets. Moreover, our proposed PolicyClusterHOGCN and PolicyClusterGCN outperform ClusterHOGCN and ClusterGCN on all the datasets. This result suggests that our approach is able to identify strong cluster configurations than prior state-of-the-art approaches. The difference in performance between PolicyClusterHOGCN and other baselines is statistically significant with a standard paired t-test at a significance level of 0.05. Another observation is that with HOGCN, the performance of ClusterHOGCN is close to or often better than that of GraphSAINT. Note that our proposed framework is not limited to GCN or HOGCN and one can experiment with other graph convolutional networks design \citep{you2020design} for further improvement in the supervised task. 

\begin{figure}[!t]
\centering
\subfloat[Facebook]{\includegraphics[width=0.55\linewidth]{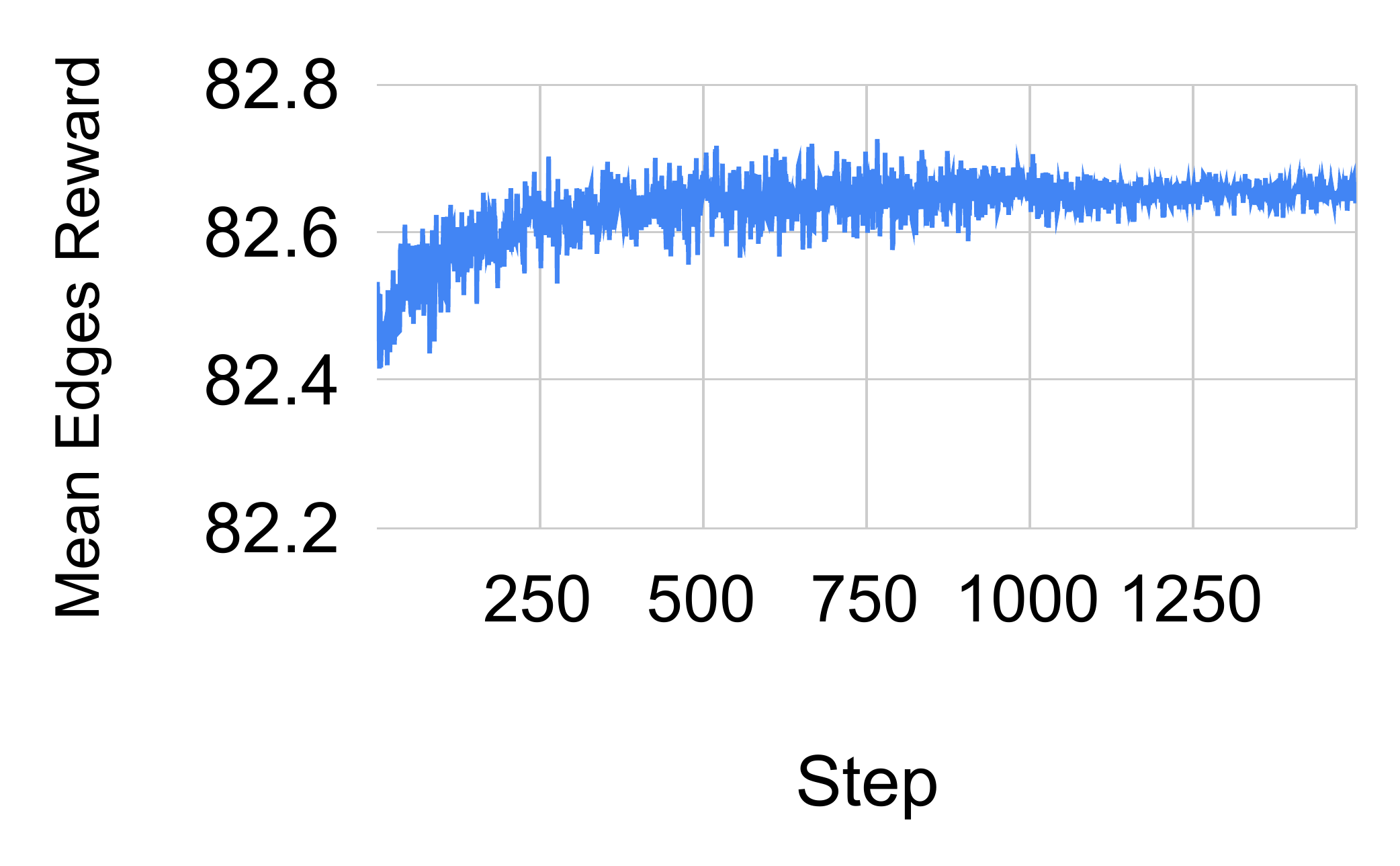}
\label{fig:rw_facebook}} 

\subfloat[Twitter]{\includegraphics[width=0.55\linewidth]{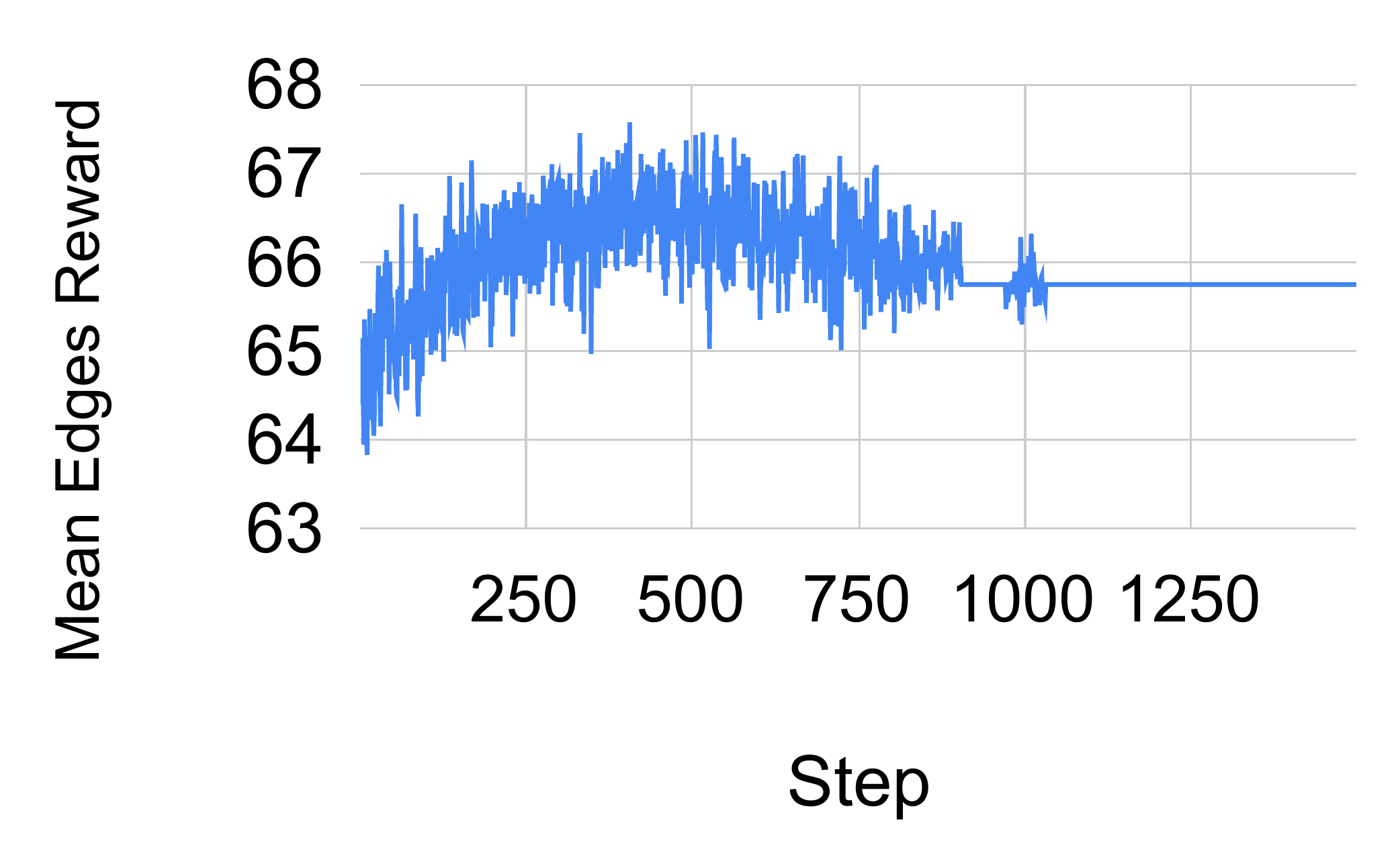}
\label{fig:rw_twitter}}
\caption{Mean edges reward per PolicyClusterHOGCN step}
\label{fig:edge_rewards}
\end{figure}

\begin{table*}[t]
\centering
    \resizebox{0.9\linewidth}{!}{
\begin{tabular}{l|l|cccccc}
\toprule
\textbf{}                            &                             & \multicolumn{1}{l}{\textbf{Croatia}} & \multicolumn{1}{l}{\textbf{Romania}} & \multicolumn{1}{l}{\textbf{Hungary}} & \textbf{Facebook}         & \textbf{Twitter}             & \textbf{Blogcatalog}      \\
\midrule

\multirow{3}{*}{PolicyClusterHOGCN} & Node2vec & 0.463 & 0.478 & 0.497 & 0.563 & 0.189 & 0.956 \\
& M-NMF   & 0.461 & 0.476 & 0.497 & 0.556 & 0.163 &  0.956\\
& Net-MF   & 0.462 & 0.478 & 0.498 & 0.562 & 0.168 & 0.956 \\
                                     
\bottomrule
\end{tabular}
}
\caption{Robustness: state embedding methods.}
\label{tab:state_emb}
\end{table*}

\begin{table*}[t]
\centering
    \resizebox{0.9\linewidth}{!}{
\begin{tabular}{l|l|cccccc}
\toprule
\textbf{}                            &                             & \multicolumn{1}{l}{\textbf{Croatia}} & \multicolumn{1}{l}{\textbf{Romania}} & \multicolumn{1}{l}{\textbf{Hungary}} & \textbf{Facebook}         & \textbf{Twitter}             & \textbf{Blogcatalog}      \\
\midrule

\multirow{3}{*}{PolicyClusterHOGCN} & Actor-critic & 0.463 & 0.478 & 0.497 & 0.563 & 0.189 & 0.956\\
      & Reinforce   & 0.461 & 0.479 & 0.488 & 0.562 & 0.177 & 0.955 \\

\bottomrule
\end{tabular}
}
\caption{Robustness: policy gradient methods.}
\label{tab:policy}
\end{table*}

\subsection{Reward Analysis}
Figure \ref{fig:edge_rewards} shows the mean of mean edge rewards received over five independent runs. We observe that as the policy network's training progresses, PolicyClusterGCN can identify better cluster configurations that result in the improvement in GCN training performance.

\subsection{Robustness: State Embedding Methods}
We test the sensitivity of the proposed PolicyClusterGCN framework with respect to the choice of state embedding methods. We consider three UGRL methods: a random-walk based method, node2vec \citep{grover2016node2vec}, a community aware embedding method, M-NMF \citep{wang2017community}, and a matrix factorization based approach, NetMF \citep{qiu2018network}. The average results over five independent runs are presented in Table \ref{tab:state_emb}. We observe that PolicyClusterHOGCN achieves similar performance on all of the six real-world datasets and is robust to the choice of state embedding method.

\begin{table}[t]
\centering
\resizebox{1.0\linewidth}{!}{
\begin{tabular}{rccc}
\toprule
\textbf{LFR$_\mu$} & \textbf{ClusterGCN} & \textbf{PolicyClusterGCN} & \textbf{\% diff.} \\
\midrule


0.10 & 0.711 & 0.731 & +2.00 \\
0.15 & 0.739 & 0.753 & +1.40 \\ 
0.20 & 0.552 & 0.561 & +0.90 \\
0.25 & 0.452 & 0.472 & +2.05 \\
0.30 & 0.223 & 0.237 & +1.40 \\
0.35 & 0.193 & 0.201 & +0.80 \\
0.40 & 0.135 & 0.144 & +0.90 \\


\bottomrule
\end{tabular}
}

\caption{Synthetic Dataset: LFR. LFR$_\mu$ corresponds to percentage of inter cluster links.}
\label{tab:lfr}
\end{table}

\subsection{Robustness: Policy Gradient Methods}
Next, we test the sensitivity of our framework with respect to the training algorithm. Here, we train PolicyClusterHOGCN with reinforce \citep{williams1992simple} and actor-critic \citep{konda1999actor} algorithms. Table \ref{tab:policy} shows the performance of PolicyClusterHOGCN. We observe that PolicyClusterHOGCN trained with actor-critic often outperforms PolicyClusterHOGCN trained with the reinforce algorithm. The performance improvement is likely due to the better ability of the actor-critic algorithm to reduce gradient variance \citep{sutton2018reinforcement}.

\subsection{Synthetic Datasets}
We compare the performance of PolicyClusterHOGCN and ClusterHOGCN on the synthetic LFR datasets \citep{lancichinetti2008benchmark}. Each LFR dataset consists of a set of communities and $\mu$ percentage of inter-community connections. In this section, we study how the performance of PolicyClusterHOGCN and PolicyClusterHOGCN changes as we change the value of $\mu$ from 0.1 to 0.4 with a 0.05 step size. We keep the rest of the parameters of the LFR dataset the same (number of nodes=5,000, average degree=5, min community size=50, power law exponent for the degree distribution and community size distribution to 3 and 1.5, respectively). We use cdlib library \citep{rossetti2019cdlib} for generating the datasets. 
A node's cluster is set as its class and we perform multi-class classification. For node features, we  factorize the adjacency matrix of these datasets and treat 16 left singular vectors as
node attributes.  The test Micro-f1 scores are shared in Table \ref{tab:lfr}. We observe that PolicyClusterHOGCN is consistently able to identify better clusters as compared to ClusterHOGCN for all $\mu$ values. However, as the percentage of inter-community edges increases, the performance between PolicyClusterHOGCN and ClusterHOGCN decreases. This result is not terribly surprising since at that point noise (represented by inter-cluster edges) tends to dominate the signal (represented by node clusters).

\begin{figure*}[!t]
\centering
\subfloat[Croatia]{\includegraphics[width=0.165\linewidth]{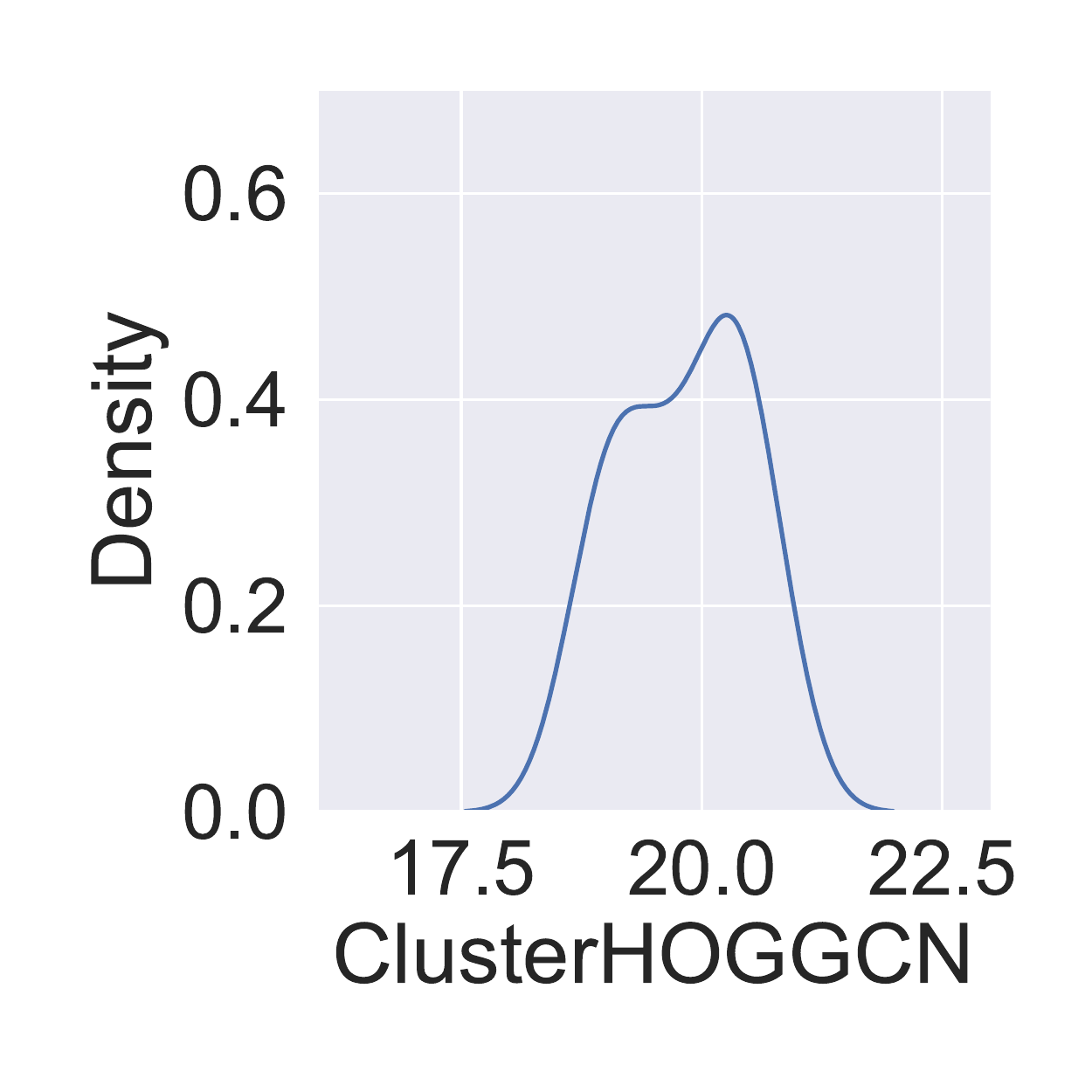}
\label{fig:CroatiaClusterGCN}}
\subfloat[Croatia]{\includegraphics[width=0.165\linewidth]{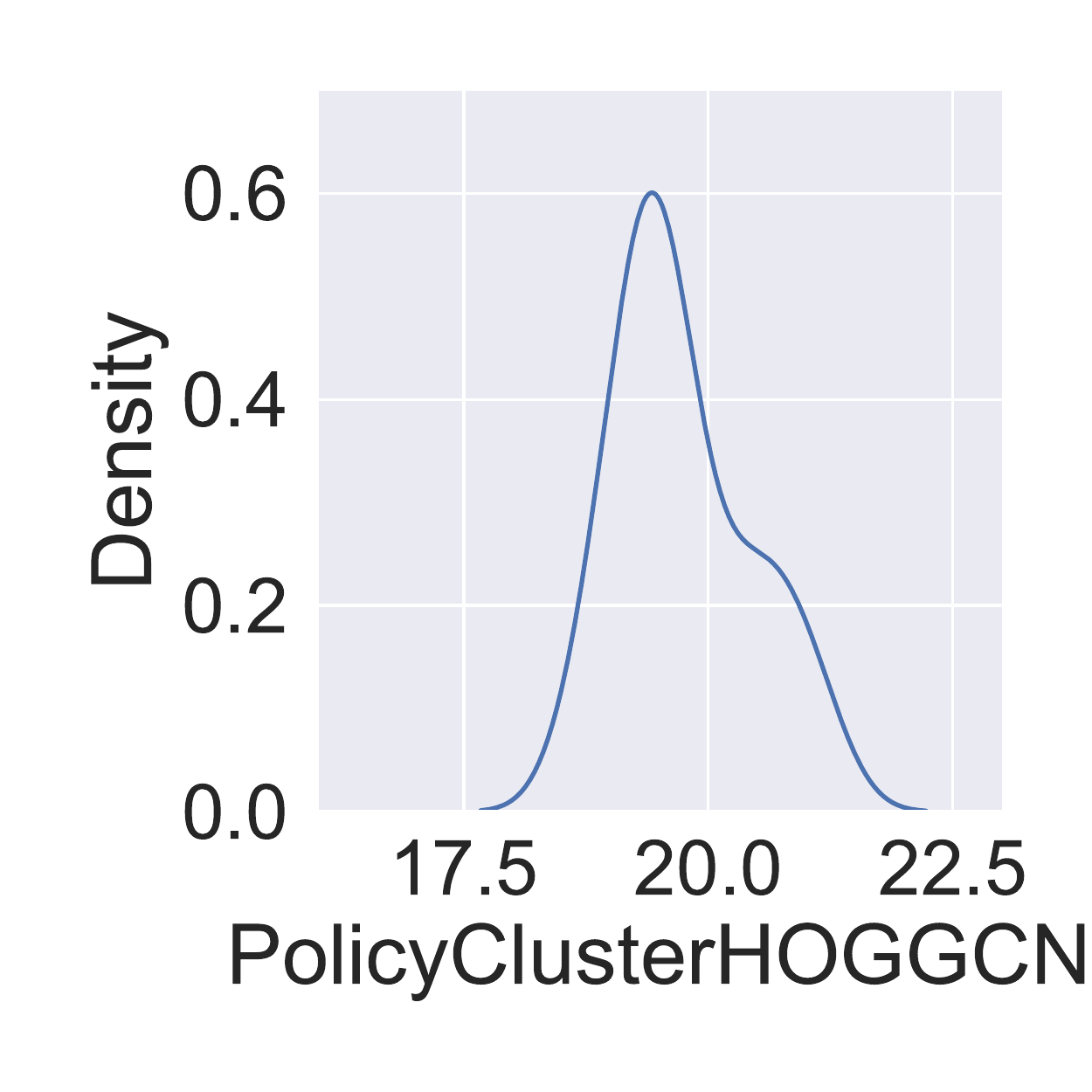}
\label{fig:CroatiaPolicyClusterGCN}} 
\subfloat[Romania]{\includegraphics[width=0.165\linewidth]{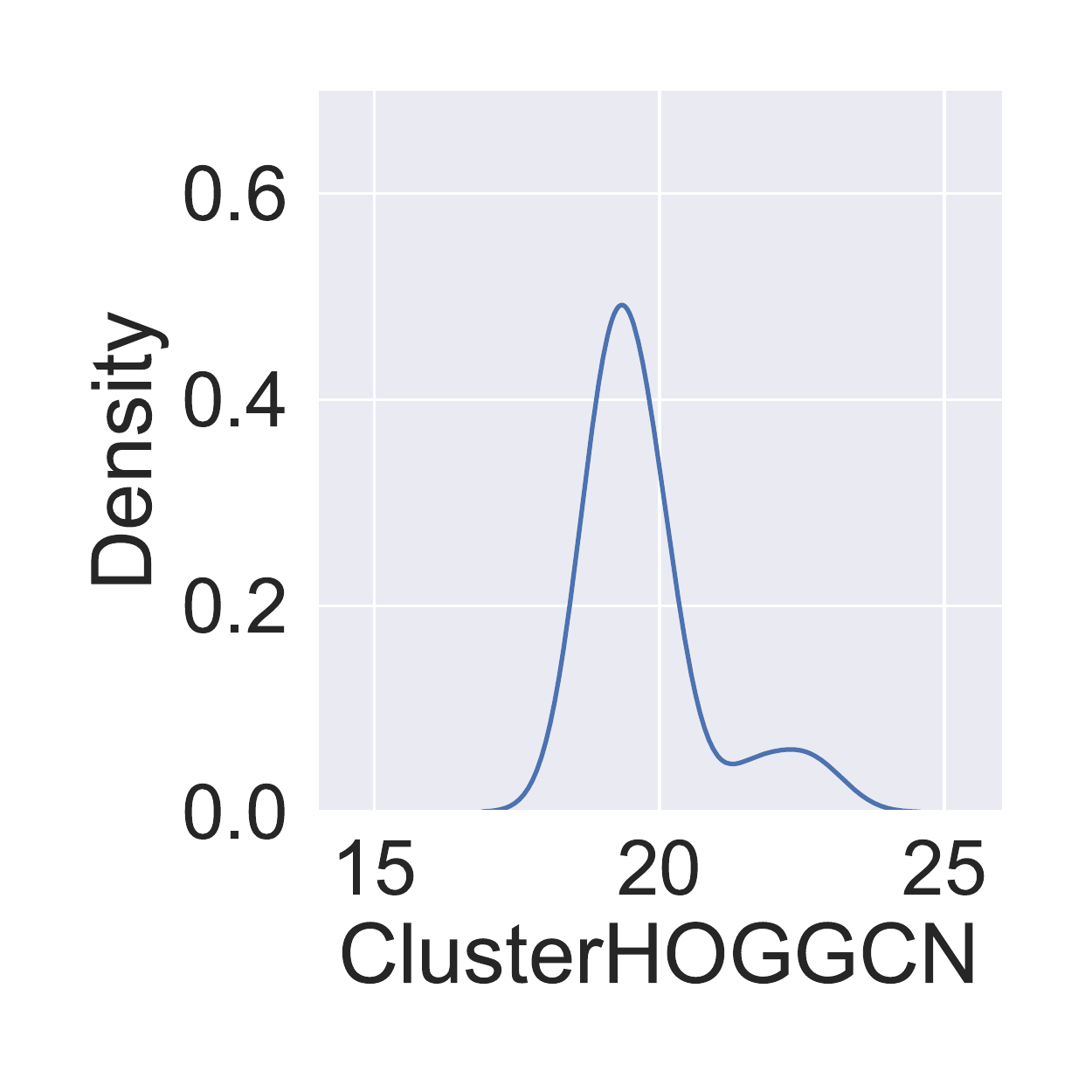}
\label{fig:RomaniaClusterGCN}}
\subfloat[Romania]{\includegraphics[width=0.165\linewidth]{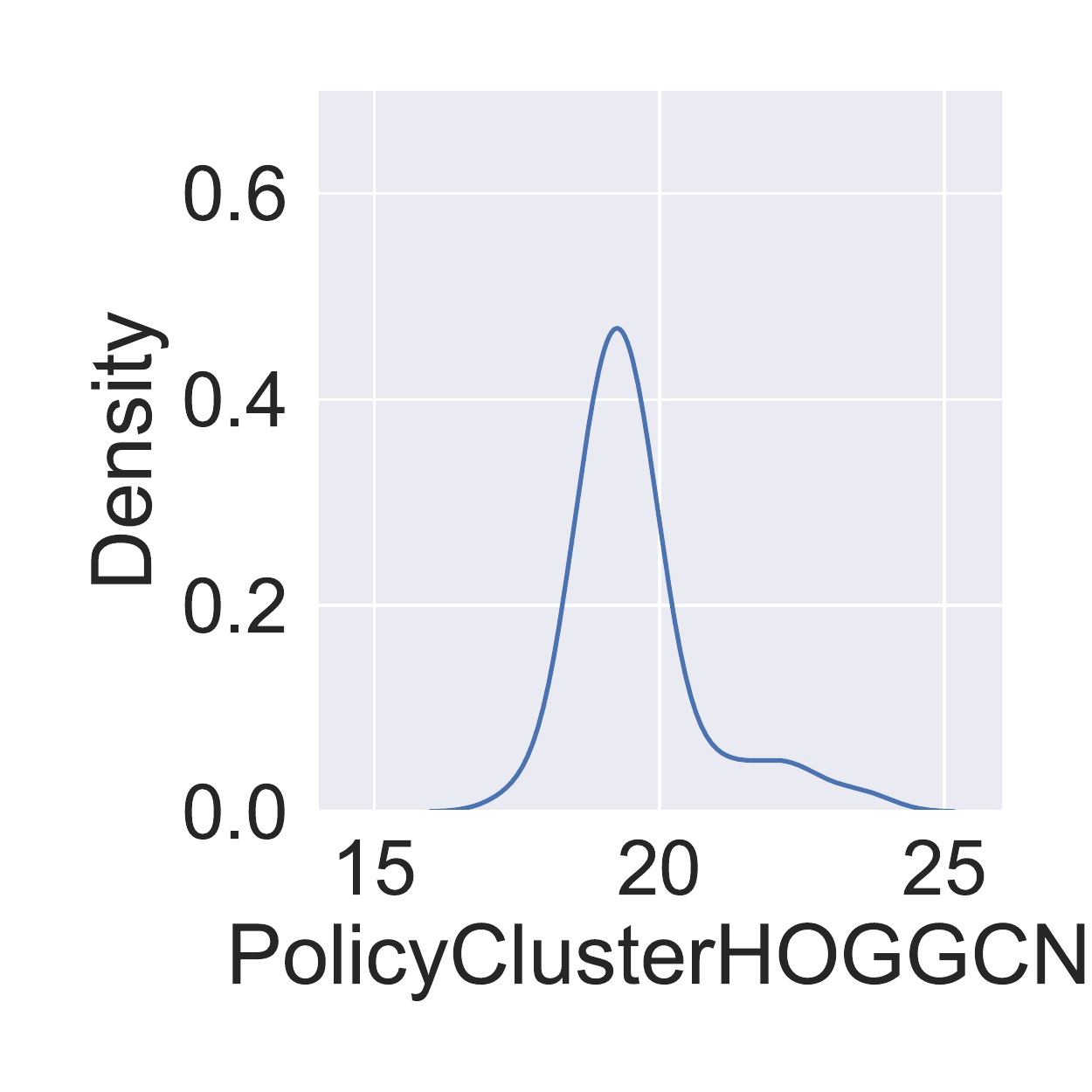}
\label{fig:RomaniaPolicyClusterGCN}} 
\subfloat[Hungary]{\includegraphics[width=0.165\linewidth]{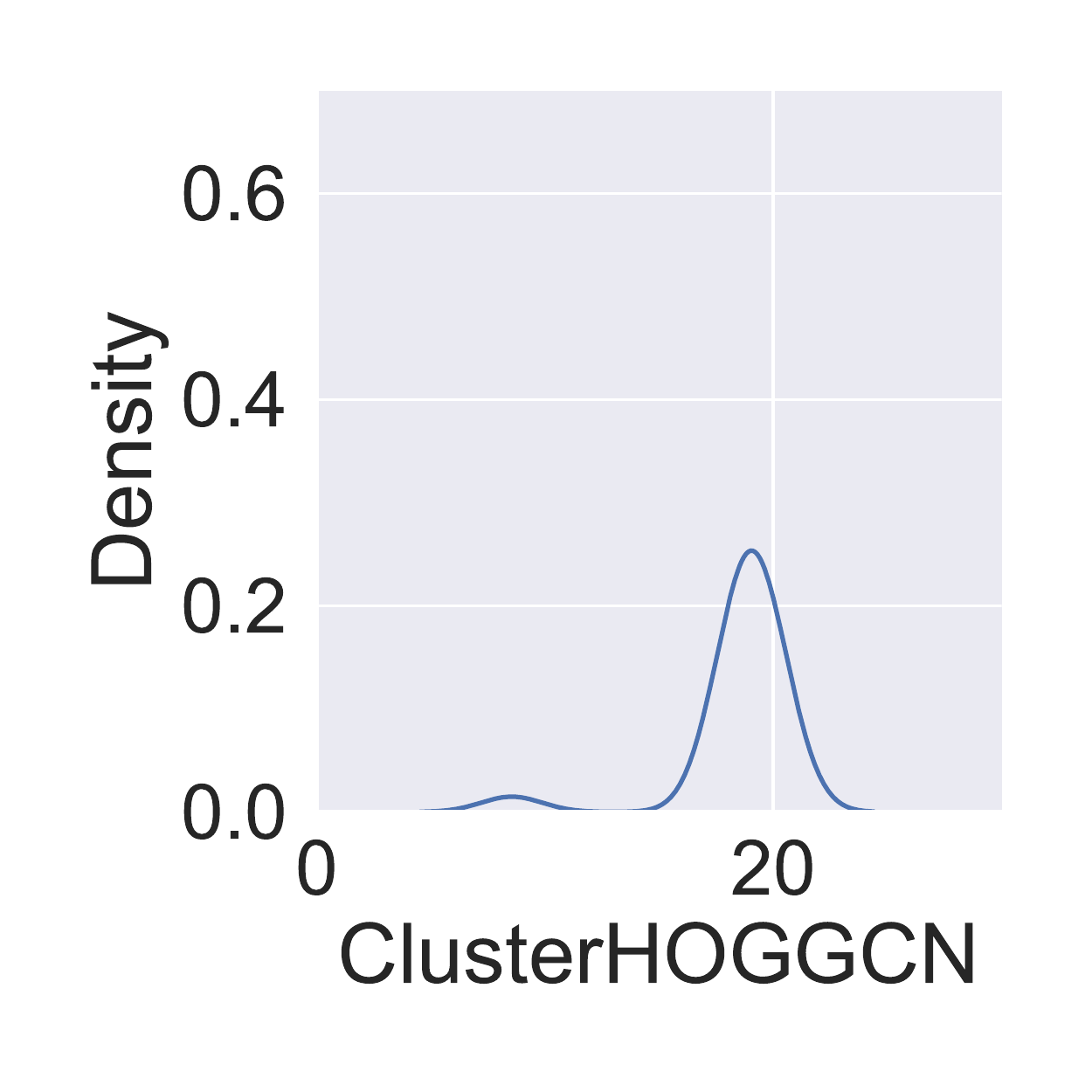}
\label{fig:HungaryClusterGCN}}
\subfloat[Hungary]{\includegraphics[width=0.165\linewidth]{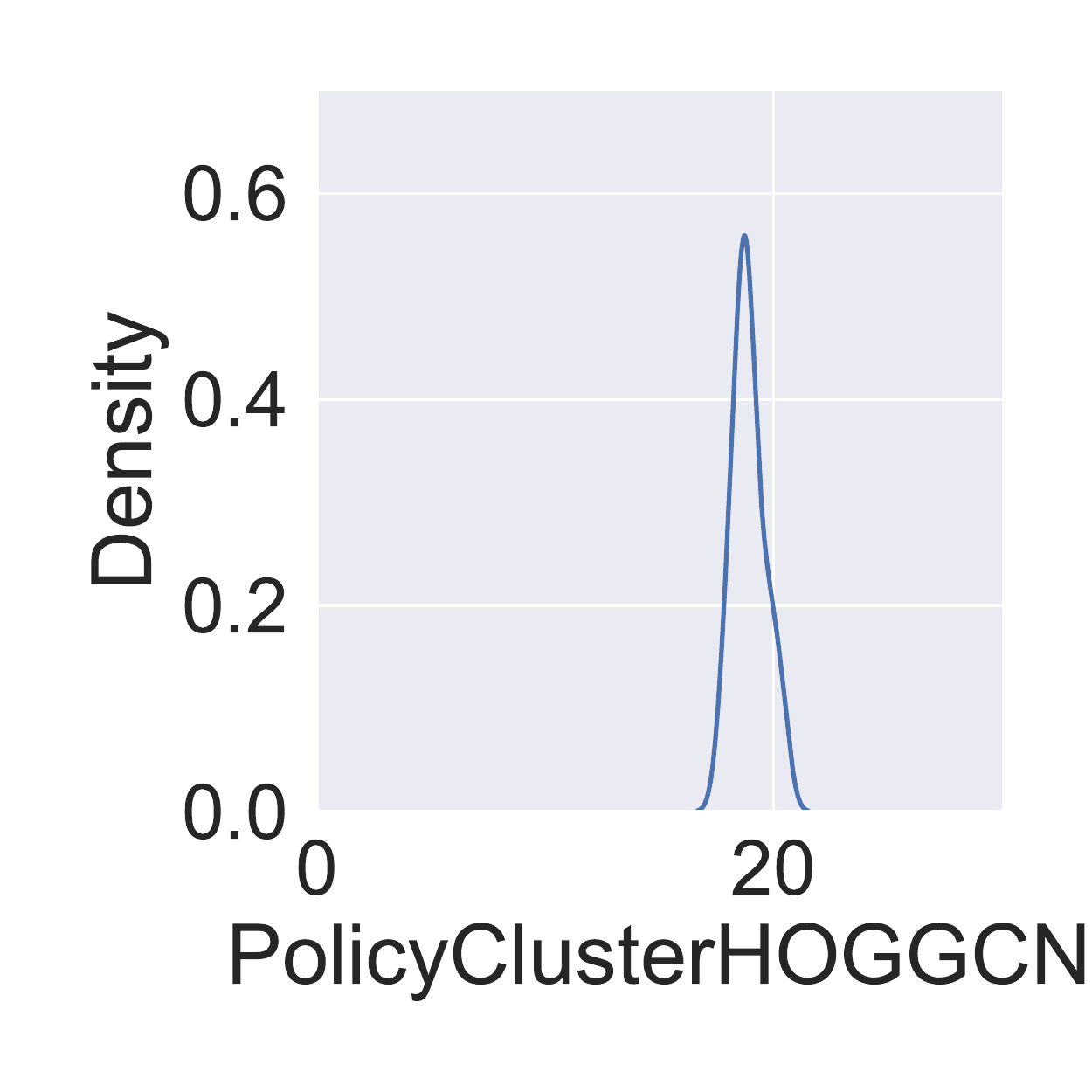}
\label{fig:HungaryPolicyClusterGCN}} \\
\subfloat[Facebook]{\includegraphics[width=0.165\linewidth]{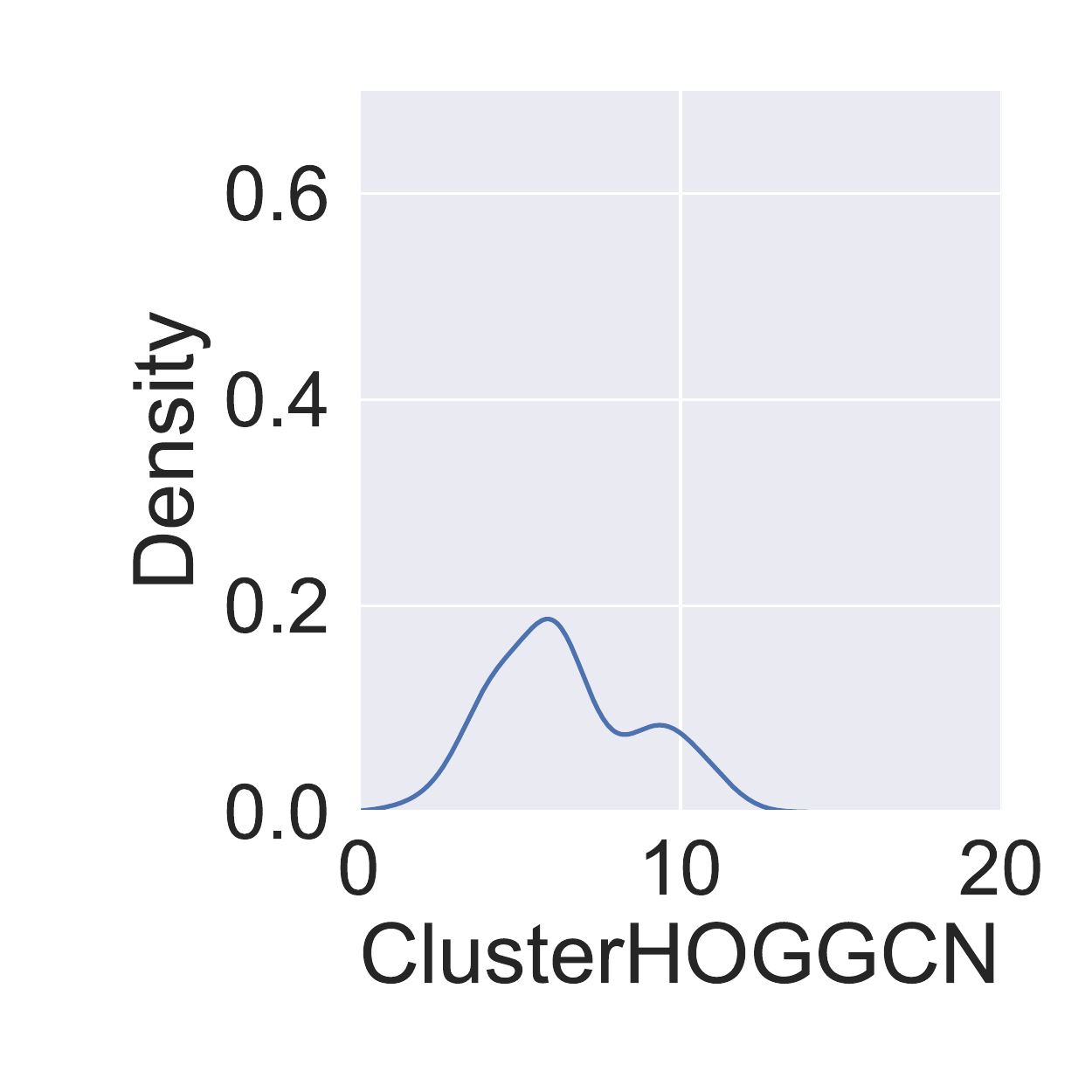}
\label{fig:FacebookClusterGCN}}
\subfloat[Facebook]{\includegraphics[width=0.165\linewidth]{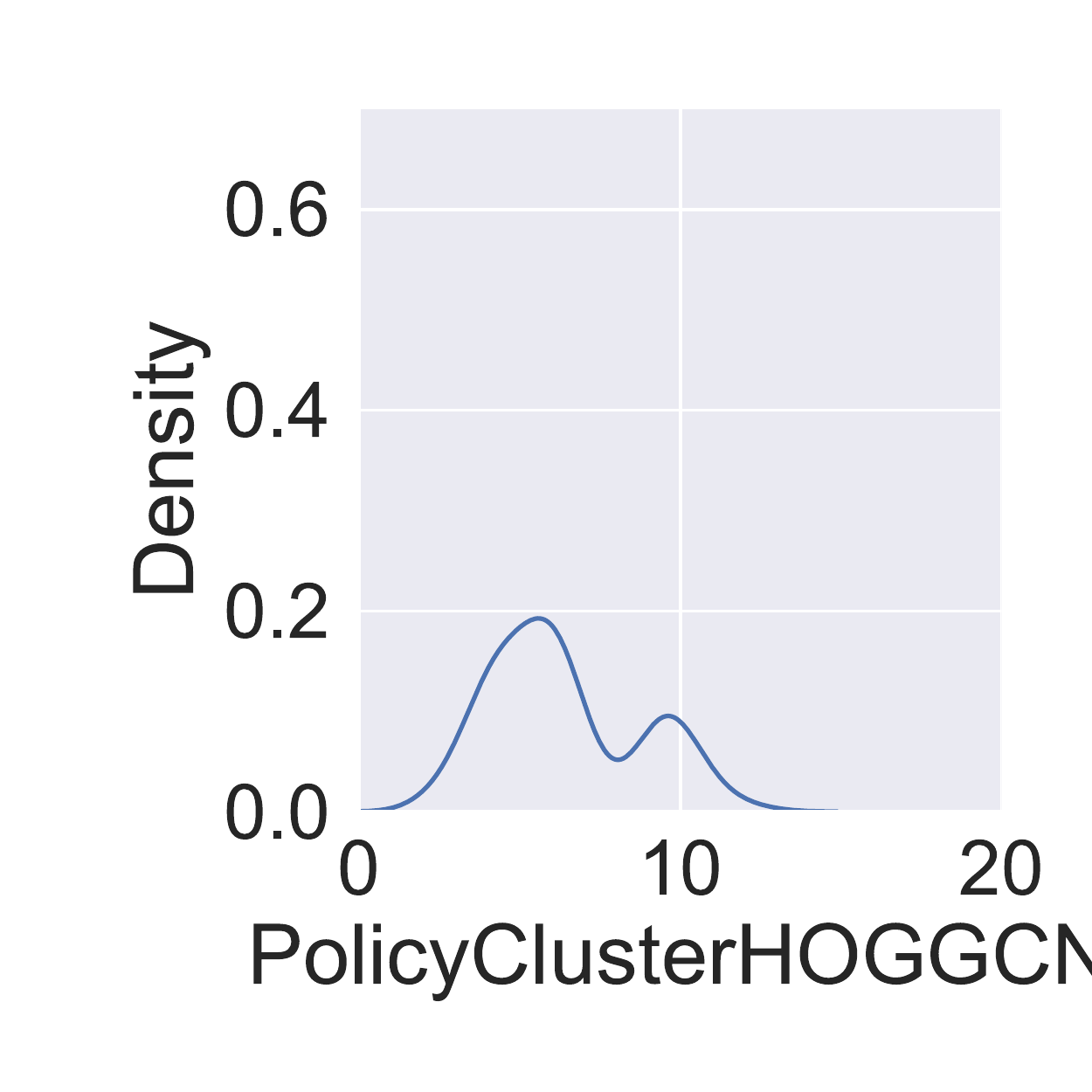}
\label{fig:FacebookPolicyClusterGCN}}  
\subfloat[Twitter]{\includegraphics[width=0.165\linewidth]{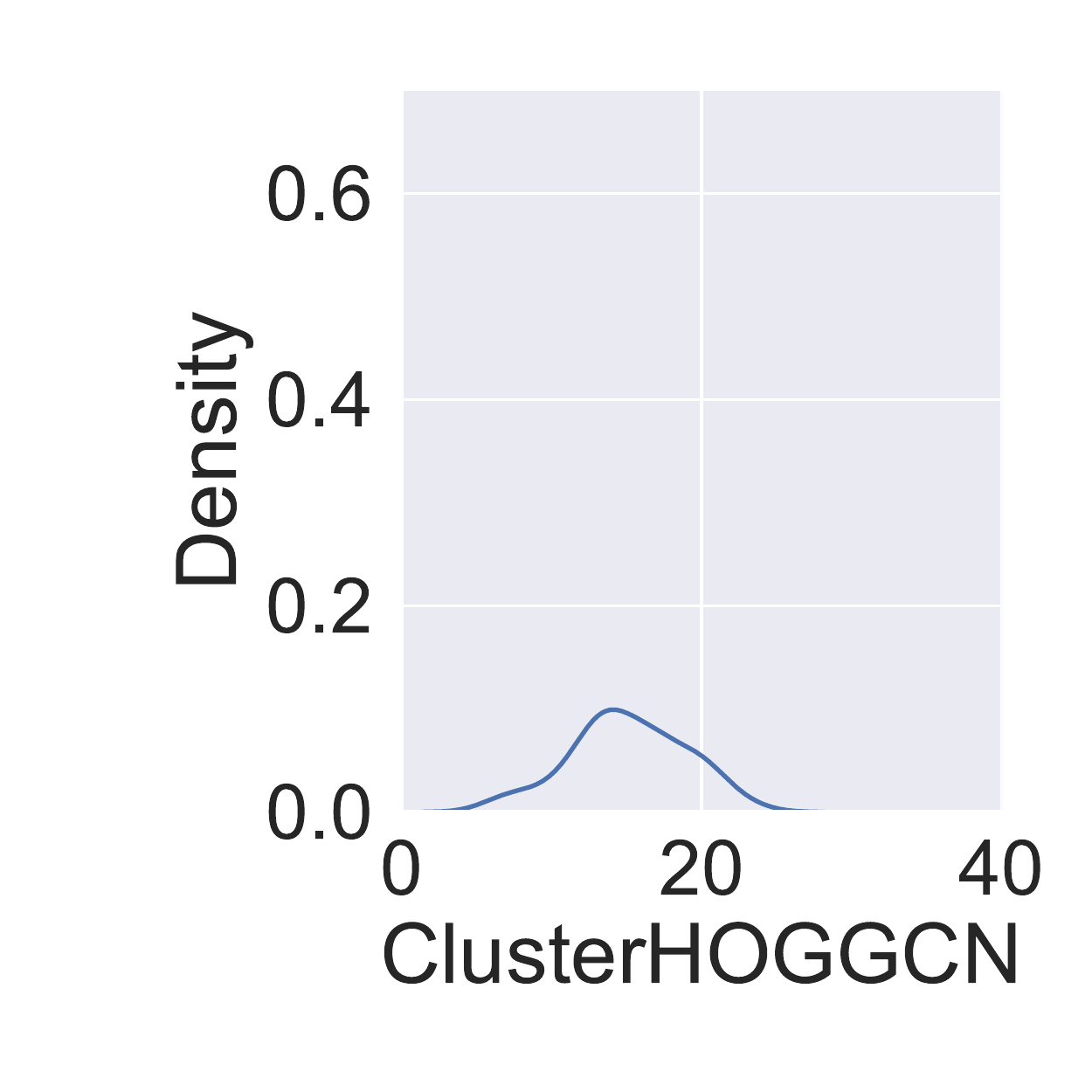}
\label{fig:TwitterClusterGCN}}
\subfloat[Twitter]{\includegraphics[width=0.165\linewidth]{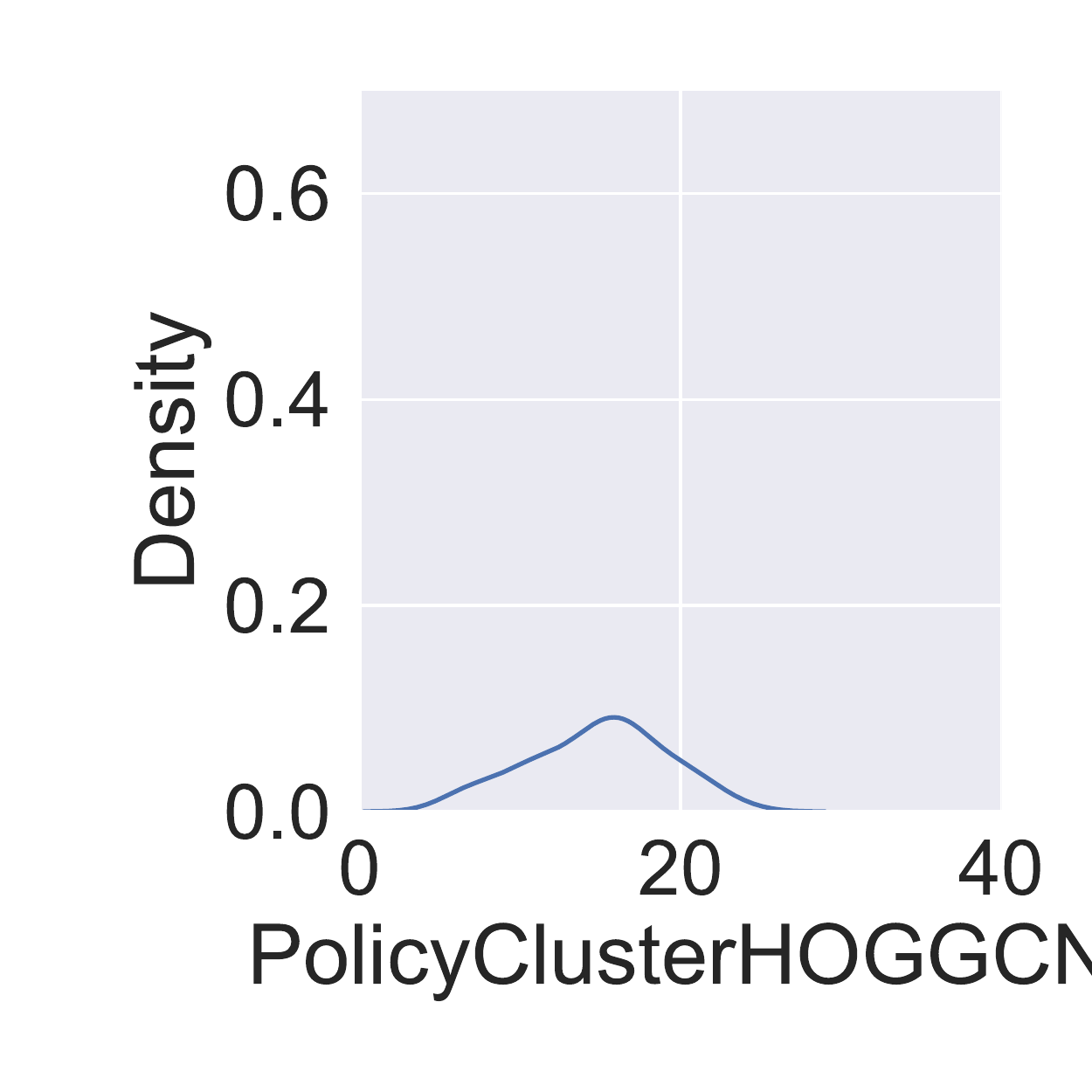}
\label{fig:TwitterPolicyClusterGCN}}
\subfloat[Blogcatalog]{\includegraphics[width=0.165\linewidth]{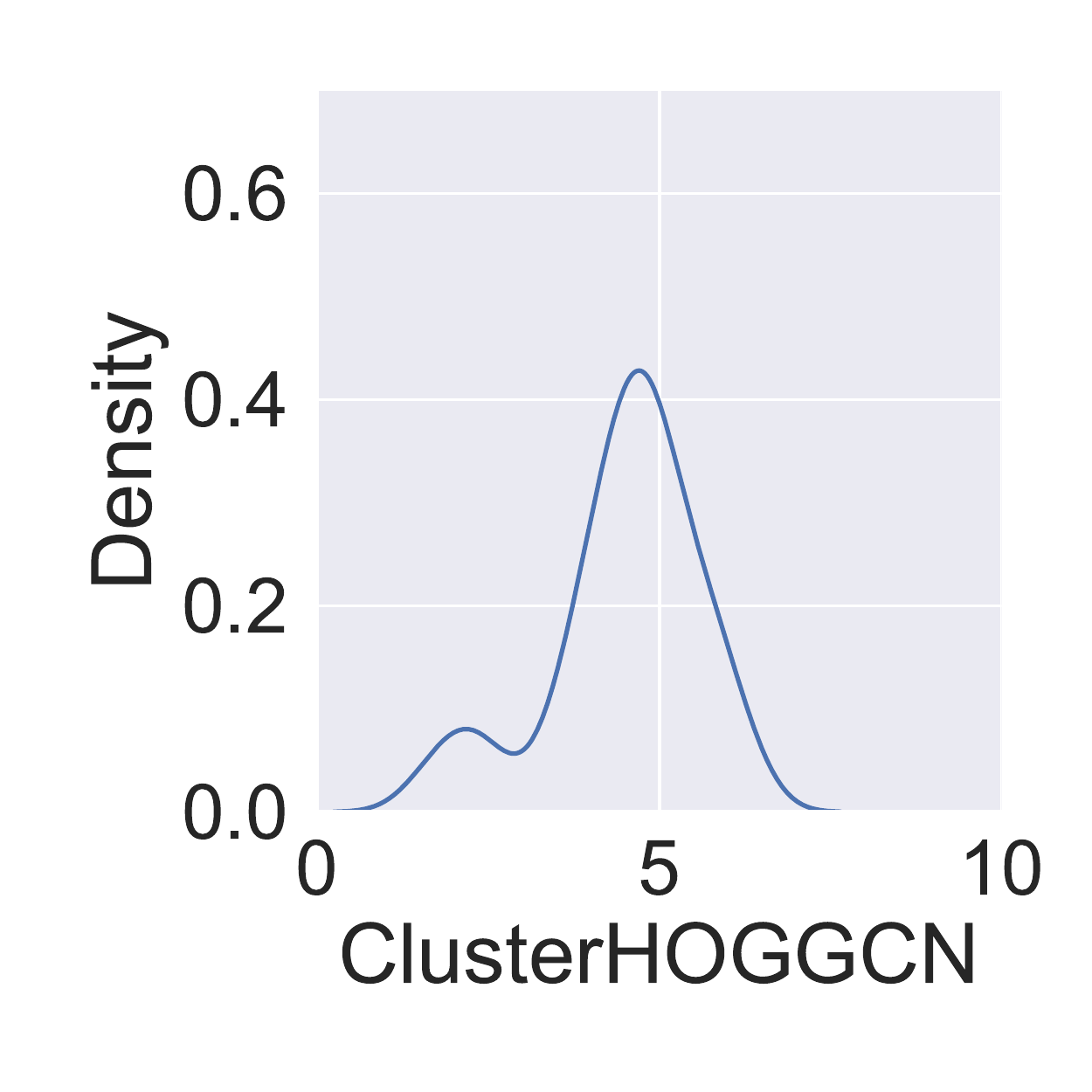}
\label{fig:BlogcatalogClusterGCN}}
\subfloat[Blogcatalog]{\includegraphics[width=0.165\linewidth]{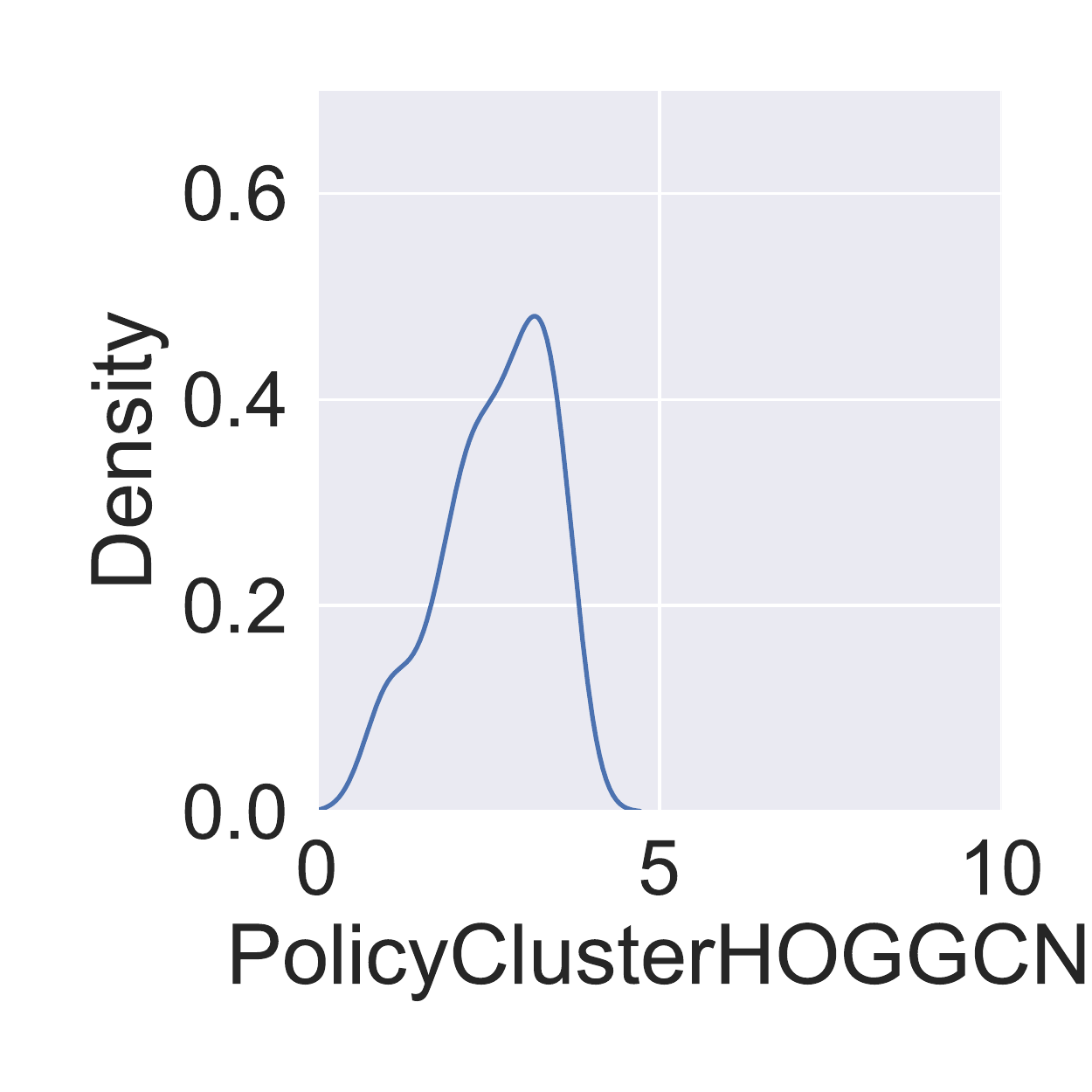} 
\label{fig:BlogcatalogPolicyClusterGCN}} \\
\caption{Label Entropy}
\label{fig:label_entropy}
\end{figure*}

\subsection{Cluster Analysis: Label Entropy}
The performance of GCN is dependent on both cluster structure and node label distribution in the cluster. Here, we compare and contrast the node labels present in the clusters identified by ClusterHOGCN and PolicyClusterHOGCN. For comparison, we rely on the label entropy metric utilized elsewhere \citep{chiang2019cluster}. To compute a cluster's label entropy, we assume labels are independent and a node's label can have 0 or 1 value. For $q$ labels, cluster $c$'s label entropy can then be calculated as $S_c = H(X_1, X_2, ..., X_q) = H(X_1)+ H(X_2)+ ...+ H(X_q)$ where $X_i$ is Bernoulli random variable. A dataset's label entropy is represented through the distribution of cluster label entropies and consists of $k$ number of clusters. Since we report the results over five independent runs, our reported distribution size is $k$ times five. Figure \ref{fig:label_entropy} presents the distribution of cluster label entropies using the kernel density estimation algorithm for six real-world datasets. We observe that, in general, the distribution of label entropies of ClusterHOGCN has high variance as compared to that of PoliyClusterHOGCN which has low variance -- see Figure \ref{fig:CroatiaClusterGCN} vs \ref{fig:CroatiaPolicyClusterGCN}, Figure \ref{fig:HungaryClusterGCN} vs \ref{fig:HungaryPolicyClusterGCN}, Figure \ref{fig:BlogcatalogClusterGCN} vs \ref{fig:BlogcatalogPolicyClusterGCN}. The low variance or high saturated label entropy's suggests that PolicyClusterGCN identifies clusters with high label uncertainty in the clusters. 

\section{Related Work}
Graph convolutional neural networks (GCNs) have shown promising results on several machine learning tasks on graph-structured data. However, GCNs require the full normalized adjacency matrix of the graph, and hence scaling them on large networks becomes challenging \citep{hamilton2017inductive, chiang2019cluster, zheng2020distdgl}. 

A plethora of efficient sampling methods has been proposed to solve the scalability bottleneck and improve the GCN training \citep{liu2021sampling}. The sampling-based techniques can be categorized as node sampling vs layer sampling vs subgraph sampling. In node sampling techniques, one usually samples a fixed number of neighbors per node and forms a subgraph consisting of multiple connected nodes and their sampled neighbors \citep{hamilton2017inductive}. VR-GCN utilizes the historical activations of nodes in GCN training to efficiently sample a much smaller number of neighbors per node. Layer sampling-based method sample usually diverse nodes per GCN layer \citep{chen2018fastgcn}. LADIES \citep{zou2019layer} improved the sampling process through importance sampling. 

Subgraph sampling-based GCN approaches have achieved state-of-the-art performances on the node classification task. These approaches often perform clustering using graph partitioning algorithms \citep{chiang2019cluster, zheng2020distdgl, zhu2019aligraph} and the identified subgraph is treated as a minibatch for GCN training. RippleWalk \citep{bai2021ripple} proposed a novel set expansion-based subgraph sampling method to construct minibatches. GraphSAINT \citep{zeng2019graphsaint} proposed several subgraph samplers to identify subgraphs and also introduced several normalization techniques to reduce the bias and variance of GCN training. 


While we have not evaluated its use for the placement of operations on computational devices we believe PolicyClusterGCN can be used for such a purpose (see Placeto \citep{addanki2019placeto}).
Placeto \citep{addanki2019placeto} learns efficient placement of node (tensorflow operations) on a compute device (GPUs). Placeto's single episode has an episode length equal to the number of nodes in the graph. One episode step places a node to a different cluster and then passes the modified graph to the environment to get the reward -- rendering the training process to be quite expensive. We believe PolicyClusterGCN can simplify this task and plan to examine such ideas in the future.  

\section*{Conclusion}


We propose PolicyClusterGCN, an online RL based approach, to identify such efficient cluster configuration for GCN training. PolicyClusterGCN's policy network modifies the edge weights of the graph that allows it to explore diverse cluster configurations. We train the policy network using an actor-critic algorithm where the reward signal is received through GCN performance. We perform experiments on six real-world datasets and show that our proposed model can outperform state-of-the-art baselines. 

In the future, we plan to explore two directions for PolicyClusterGCN: add generalizability functionality and improve scalability. To add the generalizability functionality, we propose to transform the nodes embedding of different graphs into a common embedding space and devise an efficient training algorithm for PolicyClusterGCN.
To improve scalability, we plan to explore multi-level frameworks that rely on graph coarsening to significantly reduce the size of the graph \citep{deng2019graphzoom, liang2021mile}.

\section{acknowledgements}
 This material is partially supported by the National Science Foundation (NSF) under grants
CNS-2112471, OAC-2018627, and CCF-2028944. Any opinions, fndings, and conclusions in this material are those of the author(s) and may not
refect the views of the respective funding agencies.

\bibliography{aaai23}
\end{document}